\documentclass[runningheads]{llncs}

\usepackage{eccv}

\usepackage{eccvabbrv}

\usepackage{graphicx}
\usepackage{booktabs}

\usepackage[accsupp]{axessibility}  % Improves PDF readability for those with disabilities.

\usepackage[pagebackref,breaklinks,colorlinks,citecolor=eccvblue]{hyperref}
\usepackage{orcidlink}

\usepackage{bm}
\usepackage{float}
\usepackage{graphicx}
\usepackage{caption}
\usepackage{subcaption}

\definecolor{t_p}{RGB}{79, 113, 190}
\definecolor{t_e}{RGB}{222, 131, 68}
\definecolor{t_l}{RGB}{126, 171, 85}

\usepackage{stfloats}
\usepackage{wrapfig}

\begin{document}


\title{DiffFAE: Advancing High-fidelity One-shot Facial Appearance Editing with  Space-sensitive Customization and Semantic Preservation} 

\titlerunning{DiffFAE}

\author{Qilin Wang\inst{1 \footnotemark[1]},
Jiangning Zhang\inst{2},
Chengming Xu\inst{2},
Weijian Cao\inst{2},
Ying Tai\inst{3},
Yue Han\inst{4},
Yanhao Ge\inst{5},
Hong Gu\inst{5},
Chengjie Wang\inst{2},
Yanwei Fu\inst{1 \footnotemark[4]}}

\authorrunning{Q. Wang~et al.}

\institute{Fudan University, Shanghai, China \and
Tencent Youtu Laboratory, Shanghai, China \and
Nanjing University, Nanjing, China \and
Zhejiang University, Zhejiang, China \and
VIVO, Shanghai, China}

\maketitle

\renewcommand{\thefootnote}{\fnsymbol{footnote}}
\footnotetext[1]{Work is done during the internship at Tencent YouTu Lab.} 
\footnotetext[4]{Corresponding author.}

\begin{abstract}
Facial Appearance Editing (FAE) aims to modify physical attributes, such as pose, expression and lighting, of human facial images while preserving attributes like identity and background, showing great importance in photograph. In spite of the great progress in this area, current researches generally meet three challenges: low generation fidelity, poor attribute preservation, and inefficient inference. To overcome above challenges, this paper presents DiffFAE, a one-stage and highly-efficient diffusion-based framework tailored for high-fidelity FAE. 
For high-fidelity query attributes transfer, we adopt Space-sensitive Physical Customization (SPC), which ensures the fidelity and generalization ability by utilizing rendering texture derived from 3D Morphable Model (3DMM). In order to preserve source attributes, we introduce the Region-responsive Semantic Composition (RSC). This module is guided to learn decoupled source-regarding features, thereby better preserving the identity and alleviating artifacts from non-facial attributes such as hair, clothes, and background. 
We further introduce a consistency regularization for our pipeline to enhance editing controllability by leveraging prior knowledge in the attention matrices of diffusion model. Extensive experiments demonstrate the superiority of DiffFAE over existing methods, achieving state-of-the-art performance in facial appearance editing.
\end{abstract}

\section{Introduction}
\label{sec:intro}

Facial Appearance Editing (FAE) aims to edit a source image such that physical attributes (\ie, pose, expression and lighting) from a query image can be smoothly transferred on the source identity.  The attributes such as identity, clothes, and background provided by the source image should remain consistent before and after the editing process. High-quality FAE models can be directly engaged in real-life applications among photography and social media, thus serving as a crucial tool for enhancing the aesthetic appeal of facial images.

\begin{figure}[t]
    \centering
    %\captionsetup{type=figure}
    \includegraphics[width=\textwidth]{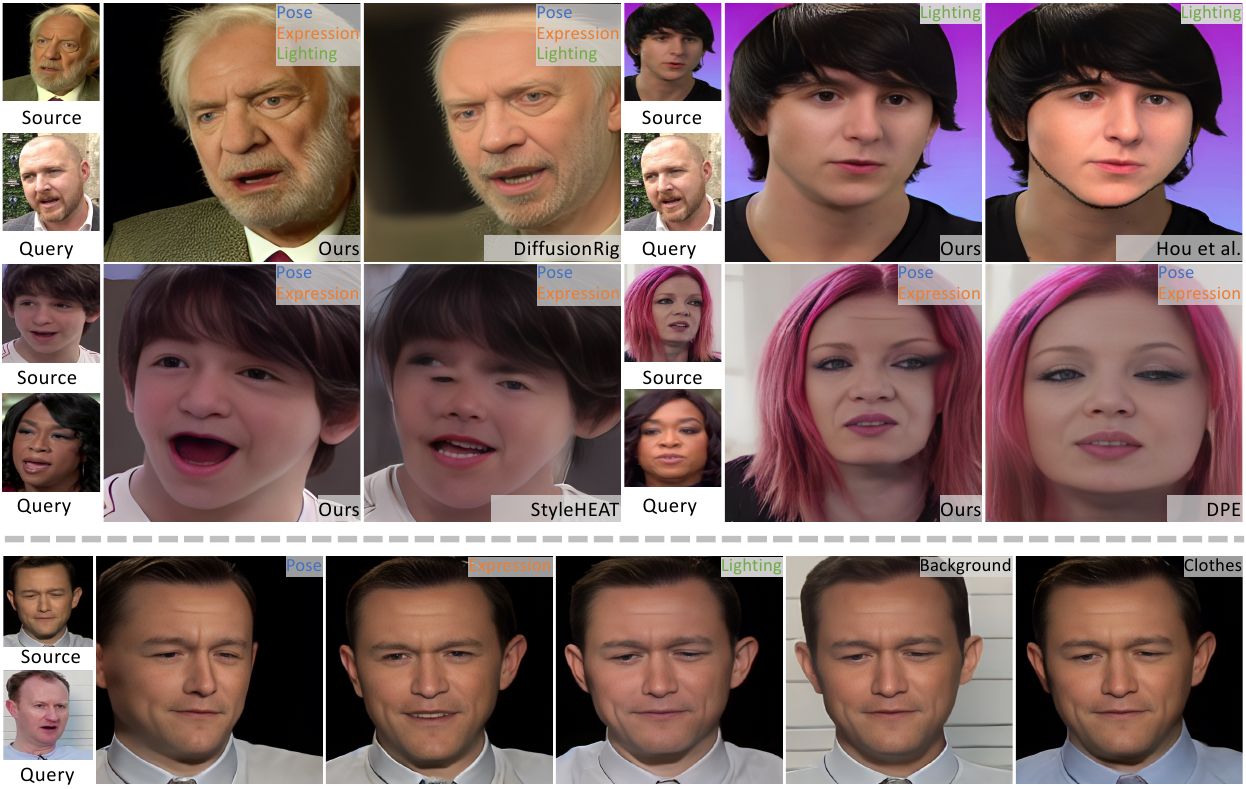}
    \vspace{-1.5em}
    \begin{picture}(0,0)
        \put(-41,160){\fontsize{5}{10}\selectfont~\cite{DiffusionRig}}
        \put(-39,87){\fontsize{5}{10}\selectfont~\cite{StyleHEAT}}
        \put(135,160){\fontsize{5}{10}\selectfont~\cite{Hou}}
        \put(149,87){\fontsize{5}{10}\selectfont~\cite{DPE}}
    \end{picture}
    \captionof{figure}{
    \textbf{Top:} Our DiffFAE produces high-fidelity facial compositional editing of three physical appearance attributes, \ie, \textcolor{t_p}{\textbf{pose}}, \textcolor{t_e}{\textbf{expression}}, and \textcolor{t_l}{\textbf{lighting}}, while achieving stronger unedited attributes preservation ability. 
    \textbf{Bottom:} DiffFAE possesses strong disentangled editing capabilities and can be easily extended to other attribute editing, such as \textbf{background} and \textbf{clothes}.
    % Compared with previous methods, our proposed DiffFAE can not only produce high-fidelity facial images with one-shot samples when editing physical attributes such as pose, expression and lighting, but also be extended to other editable attributes.
    }
    \vspace{-2.5em}
    \label{fig:teaser}
\end{figure}
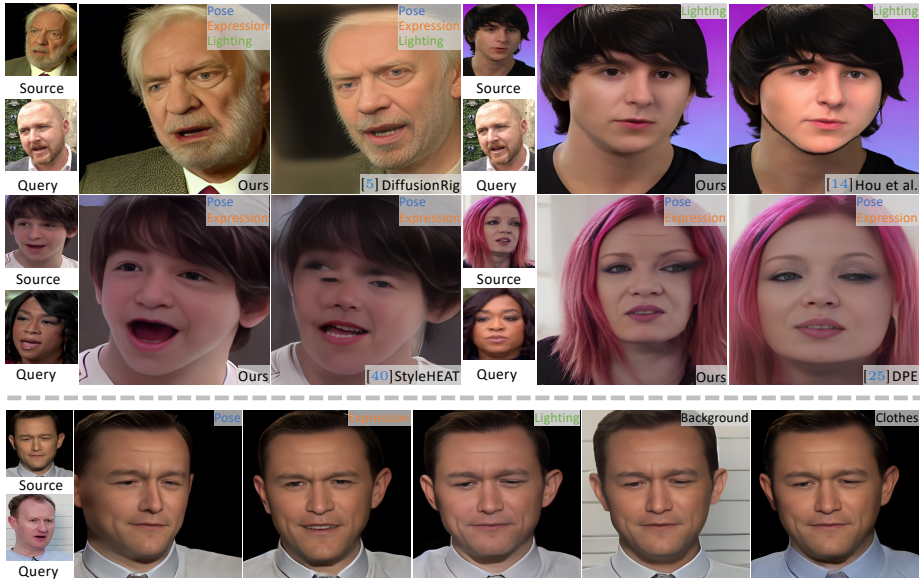

With the surge of generative models like GANs and diffusion models, several works~\cite{StyleRig,HeadGAN,PIRenderer,StyleHEAT,LIA,DaGAN,DiFaReli,Hou,DiffusionRig,MyStyle} have been proposed to solve FAE based on these powerful base models. In spite of the good performance as claimed by these works, we find that they suffer from three main challenges: (1) \textbf{Low generation fidelity}: The successful usage of FAE models must depend on high-fidelity generation results. However, current methods, especially GAN-based StyleHEAT~\cite{StyleHEAT} and PIRenderer~\cite{PIRenderer}, can frequently produce undesirable results, containing insufficient details or too much distortion. This may be attributed to the limited capacity of GANs, which makes these models unable to learn the complex information contained in both source and query images. (2) \textbf{Poor attribute preservation}: Compared with the GANs mentioned before, methods utilizing diffusion models like DiffusionRig~\cite{DiffusionRig} generally enjoy better fidelity. Nonetheless, such methods, along with GAN-based ones, typically take the source image as a whole for editing, yet fail to decompose those attributes to be preserved. Consequently, the prior knowledge rather than source information dominates in these methods for generating the source-related regions, thus leading to obvious artifacts and unsatisfactory performance. (3) \textbf{Inefficient inference}: Some FAE methods like DiffusionRig and MyStyle~\cite{MyStyle} adopt two-stage pipeline for better quality. Typically, a generalized model is first trained. Then for each person, they further finetune the pretrained model with other data collected for the specific person. Unfortunately, the data scarcity problem can easily weaken such pipelines. Even though there is way to collect the finetune data, the finetune process can take more time than expected by an application user. Therefore, the finetuning-based two-stage methods are both data and computation inefficient. 

Given such insight, it seems in great need for a ground-new FAE framework that can solve all problems mentioned above. To this end, we propose DiffFAE, a one-stage latent \textbf{Diff}usion model tailored for high-fidelity one-shot \textbf{F}acial \textbf{A}ppearance \textbf{E}diting. Our proposed DiffFAE can effectively achieve compositional and disentangled editing, as shown in \cref{fig:teaser}. Essentially, DiffFAE seperately deals with source and query information for better control. By leveraging 3D Morphable Face Model~\cite{3DMM} (3DMM), the query attributes can be actively rendered for editing. On the other hand, the source-regarding attributes are disentangled from source images in DiffFAE. In this way they can be better used as condition and preserved in the latent space.

In specific, we build our DiffFAE based on latent diffusion model (LDM)~\cite{LDM}, which has been known to enjoy powerful image generation ability in different tasks. Our model mainly relies on two modules for processing source and query condition. First, we adopt the Space-sensitive Physical Customization (SPC) module to deal with query attributes. The DECA model~\cite{DECA} is used to estimate the target parameters required by the 3DMM, specifically the FLAME model~\cite{FLAME}. These parameters are then rendered as the query texture image, which can provide sufficient query information via being used together with latent noise as the input of diffusion model. For preserving source-regarding attributes, inspired by LDM, we aim to obtain conditional feature representations with disentangled and semantically meaningful visual prompts. To achieve this, the Region-responsive Semantic Composition module (RSC) is proposed to decouple human face features into semantic visual tokens. Specifically, the identity token is extracted by ArcFace~\cite{ArcFace}, which is then used to control the denoising process via AdaIN~\cite{huang2017adain}. For the other attributes, we introduce the slot attention mechanism~\cite{SlotAttention} to DiffFAE. Through unsupervised learning, the semantic tokens learn to explicitly parse portrait features into four separate components, namely face, hair, clothes, and background. Compared with directly using global source image features as conditional information, the semantic tokens can separately represent different semantic areas in each image and influence corresponding parts of the generated image, which plays a similar role as textual tokens in LDM. Such properties can further facilitate expandable attributes editing, such as changing the background or clothes, which can hardly be performed by previous methods. Moreover, to enhance the attributes preservation, we introduce an attention consistency regularization, in which the gap between cross attention maps from each layer of LDM and slot attention maps are bridged, hence better restricting the diffusion model with prior knowledge learned by the slot features.

Extensive experiments demonstrate that DiffFAE achieves state-of-the-art performance in terms of \textbf{better generation fidelity}, \textbf{stronger attributes preservation} and \textbf{higher efficiency} compared to other existing methods. In summary, we make the contributions as follows:
\begin{itemize}

    \item We innovatively introduce a one-stage and highly efficient latent diffusion model specifically designed for the generation of high-fidelity controllable portraits. Especially, this model eliminates the need for test-time finetuning, streamlining the facial appearance editing process.
    
    \item To better preserve the source-regarding attributes, we propose  novel Region-responsive Semantic Composition (RSC) module to learn key information from source image respectively, which not only helps improve generation quality but also benefits expandable editing.

    \item We propose a novel attention consistency regularization to guide the model to focus on corresponding semantic regions of human faces.

\end{itemize}

\section{Related Work}
\label{sec:relatedwork}

\noindent\textbf{Conditional Image Generation.} Conditional image generation has long been considered as one of the most important problems for generative models. Previous methods can be categorized by types of generative models, \ie, GAN-based methods like StyleGAN~\cite{StyleGAN}, BigGAN~\cite{brock2018biggan}, auto-regression methods like VQGAN~\cite{esser2021vqgan} and DALLE~\cite{ramesh2021dalle}. Recently the researchers have been concentrated on diffusion models, which generally show better quality than GANs and auto-regressive models. For example, Rombarch~\etal proposed Latent Diffusion Model (LDM) \cite{LDM} in which the latent noise is guided by textual features predicted by CLIP \cite{radford2021clip}. Following works such as DiT~\cite{li2022dit}, PIXART-$\alpha$~\cite{chen2023pixart} and Consistency Model~\cite{song2023consistency} try to improve LDM with different structures. On the other hand, some works try to control the diffusion model with other modalities of conditional information. For example, ControlNet~\cite{zhang2023controlnet} introduces a simple additional module which can produce various kinds of conditions. Zheng~\etal~\cite{zheng2023layoutdiffusion} propose to leverage layout information in the denoising process, which can control the spatial relation between different objects in each image. We in this paper propose to consider the FAE problem as a image-conditioned image generation problem using diffusion models. Unlike the existing methods, we introduce a novel way to deal with conditional image by leveraging slot attention mechanism, which can benefit the model in terms of source attribute preservation.

\noindent\textbf{Facial Appearance Editing.} Facial Appearance Editing (FAE) aims to edit the source human face based on guidance attributes provided by a query image. Methods for this task mainly utilize 3D prior knowledge. For example, the GAN-based StyleRig~\cite{StyleRig} leverages 3DMM to edit faces from a pre-trained StyleGAN. HeadGAN~\cite{HeadGAN} proposes to extract and adapt 3D head representations from driving videos. PIRenderer~\cite{PIRenderer} explores controlling face motion with 3DMM parameters. DiFaReli~\cite{DiFaReli} edits lighting with a rendered shading reference but cannot modify pose and expression. Recently, DiffusionRig~\cite{DiffusionRig} has proposed to build the FAE model based on diffusion models. While such methods have greatly driven the progress in the area, we mainly observe that there are not perfect solutions to the primary challenges as raised in Sec.~\ref{sec:intro}, \ie, low generation fidelity, poor attributes preservation and inefficient inference. To this end, we propose a novel pipeline for FAE, by which these three problems can be well solved.

\noindent\textbf{Object-centric Learning.}
This technique is originated for tasks like object discovery and set prediction, where models learn representations of each object in a complex scene. Slot attention~\cite{SlotAttention} propagates patch-level image features to slot features using an attention mechanism, showing desirable parsing properties at both object and semantic levels. Recent improvements, like SLATE~\cite{SLATE} and STEVE~\cite{STEVE}, use transformer-based decoders, while SlotDiffusion\cite{SlotDiffusion} and LSD~\cite{LSD} propose diffusion-based decoders, significantly enhancing expressive capabilities. However, these methods primarily focus on finding feature representations for individual objects in a scene. We apply this concept to explore object-centric learning in the context of facial images.

\section{Preliminaries}

\noindent\textbf{Latent Diffusion Model (LDM).} Diffusion models~\cite{DDPM,DDIM} are probabilistic models that learn a data distribution, denoted as $p_\theta\left(\boldsymbol{x}_0\right)$. This distribution is obtained by progressively denoising a standard Gaussian distribution through a process represented as $p_\theta\left(\boldsymbol{x}_0\right)=$ $\int p_\theta\left(\boldsymbol{x}_{0: T}\right) d \boldsymbol{x}_{1: T}$, where $\boldsymbol{x}_{1: T}$ represents intermediate denoising results. The forward process of diffusion models is a Markov Chain that iteratively adds Gaussian noise to the clean data $\boldsymbol{x}_0$.
During training, a denoising model $\epsilon_\theta\left(\boldsymbol{x}_t, t\right)$ is trained to predict the noise applied to a noisy sample. However, training directly in pixel space is time-consuming. With pre-trained perceptual compression models consisting of $\mathcal{E}$ and $\mathcal{D}$, Latent Diffusion Model~\cite{LDM} (LDM) can compress image $x$ from pixel space into latent vector $z$ in the latent space, which allows the model to focus more on the semantic information of data and improves efficiency. The training objective of LDM is as follows:
\begin{equation}
\label{eq:ldm}
\mathcal{L}_{LDM}=\mathbb{E}_{\mathcal{E}(x), \epsilon \sim \mathcal{N}(0,1), t}\left[\left\|\epsilon-\epsilon_{\theta}\left(z_t, t\right)\right\|_2^2\right],
\end{equation}
where $t$ is uniformly sampled from $\{1,...,T\}$.

\noindent\textbf{3D Morphable Model (3DMM).}
3DMM~\cite{3DMM} is a type of model used to represent 3D face, utilizing a parametric space with explicit semantic information. FLAME~\cite{FLAME} is one of such models that is employed in our work. FLAME generates a mesh comprising 5,023 vertices based on the input parameters of face shape $\boldsymbol{\beta} \in \mathbb{R}^{|\boldsymbol{\beta}|}$, pose $\boldsymbol{\rho} \in \mathbb{R}^{3 k+3}$ (with $k=4$ joints for neck, jaw, and eyeballs), and expression $\boldsymbol{\psi} \in \mathbb{R}^{|\psi|}$. The model is defined as:
\begin{equation}
\label{eq:3dmm}
    M(\boldsymbol{\beta}, \boldsymbol{\rho}, \boldsymbol{\psi})=W\left(T_P(\boldsymbol{\beta}, \boldsymbol{\rho}, \boldsymbol{\psi}), \mathbf{J}(\boldsymbol{\beta}), \boldsymbol{\rho}, \mathcal{W}\right),
\end{equation}
where $W(\mathbf{T}, \mathbf{J}, \boldsymbol{\rho}, \mathcal{W})$ represents the blend skinning function that rotates the vertices in $\mathbf{T} \in \mathbb{R}^{3 n}$ around joints $\mathbf{J} \in \mathbb{R}^{3 k}$, smoothly influenced by blendweights $\mathcal{W} \in \mathbb{R}^{k \times n}$. $T_{P}$ denotes the template with added shape, pose, and expression offsets. The joint locations $\mathbf{J}$ are defined as a function of the shape $\boldsymbol{\beta}$. FLAME primarily provides facial geometry information while lacks facial texture information. Therefore, we employ DECA~\cite{DECA} in this paper, which is an off-the-shelf estimator that can predict albedo, lighting, and FLAME coefficients from a single input facial image. With these coefficients, we can obtain rendering texture representation that encompasses information about various physical attributes, \eg, pose, expression, lighting, \etc.

\section{Method}

\begin{figure}[H]
    \centering
    \vspace{-2em}
    \includegraphics[width=1.0\linewidth]{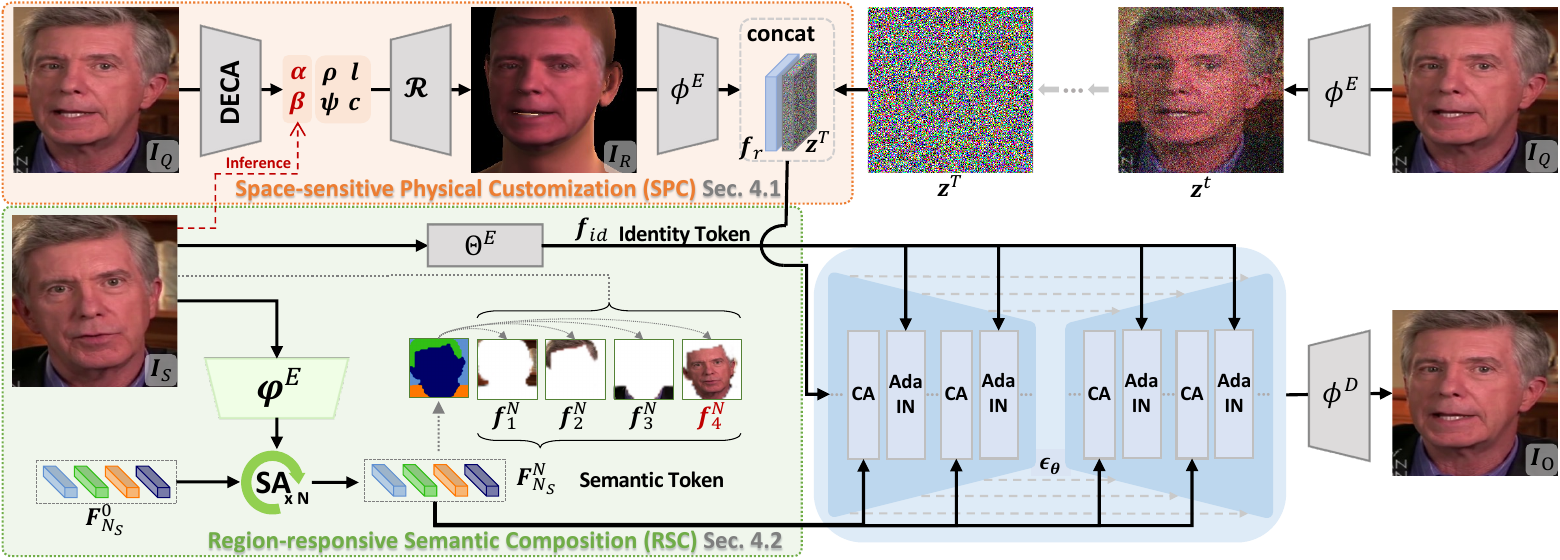}
    \caption{\textbf{Overview of the proposed DiffFAE framework}, which consists of: \textbf{\textit{1)}} \textbf{Space-sensitive Physical Customization (SPC)} takes query image $\bm{{I}}_{Q}$ as input, which goes through DECA~\cite{DECA} to extract physical coefficients, \ie, albedo $\boldsymbol{\alpha}$, shape $\boldsymbol{\beta}$, camera $\boldsymbol{c}$, pose $\boldsymbol{\rho}$, expression $\boldsymbol{\psi}$, and Spherical Harmonics lighting $\boldsymbol{l}$. These parameters are rendered by renderer $\boldsymbol{\mathcal{R}}$ to get facial texture $\bm{{I}}_{R}$, which is then compressed by pretrained VQ-VAE encoder $\bm{{\phi}}^{E}$ to get its latent representation $\bm{{f}}_{r}$. Then concatenated $\bm{{f}}_{r}$ and noisy latent code $\bm{{z}}^{T}$ are viewed as physical attributes conditioning. 
    \textbf{\textit{2)}} \textbf{Region-responsive Semantic Composition (RSC)} includes a region-responsive encoder $\boldsymbol{{\varphi}}^{E}$ and a N-iteration Slot-Attention (SA) module to extract four decoupled feature vectors $\bm{{F}}_{N_{S}}^{N}=\{\bm{{f}}_{1}^{N}, \bm{{f}}_{2}^{N}, \bm{{f}}_{3}^{N}, \bm{{f}}_{4}^{N}\}$ from a randomly initialized $\bm{{F}}_{N_{S}}^{0}$, which represent four different regions from source image $\bm{{I}}_{S}$. Furthermore, an identity extractor $\bm{\Theta}^E$~\cite{ArcFace} parallelly encodes $\bm{{I}}_{S}$ into an embedding $\bm{{f}}_{id}$, which is used together with $\bm{{F}}_{N_{S}}^{N}$ to modulate the denoising U-Net $\bm{{\epsilon}}_{\theta}$ via AdaIN and cross attention, respectively. Finally, the decoder $\bm{\phi}^{D}$ transforms the output of $\bm{\epsilon}_{\theta}$ to the generated output image $\bm{{I}}_{O}$. Notably, $\bm{{I}}_{Q}$ and $\bm{{I}}_{S}$ share the same identity during training. During inference, identity-related physical attributes $\bm{{\alpha}}$/$\boldsymbol{\beta}$ and region attribute $\bm{{f}}_{4}^{N}$ are from $\bm{{I}}_{S}$, marked in \textcolor{red}{\textbf{red}}, while other attributes can be customized from any image.  
    }
    \label{fig:difffae}
    % \vspace{-1.0em}
\end{figure}

\subsection{Overview}

\noindent\textbf{Problem formulation.} Formally, Facial Appearance Editing (FAE) targets processing a pair of source and query image $\bm{I}_{S}, \bm{I}_{Q} \in \mathbb{R}^{3\times H\times W}$, where $H,W$ denotes image height and width, individually containing the source-regarding attributes such as identity, and the desired query attributes such as pose, expression and lighting. The models are required to edit $\bm{I}_{S}$ such that attributes from $\bm{I}_{Q}$ can be smoothly transferred on the source identity. 

\noindent\textbf{Model pipeline.} In order to realize high fidelity, desirable attributes preservation and high efficiency, we in this paper propose a novel and effective pipeline for this task named DiffFAE, whose overview is shown in \cref{fig:difffae}. Concretely, during training, we utilize 3DMM to extract query-specific attributes, which will be described below, resulting in the rendered texture image $\bm{I}_{R}$ representing the comprehensive physical attributes. Then the VAE encoder $\bm{\phi}^{E}$ of LDM respectively maps $\{\bm{I}_{Q}, \bm{I}_{R}\}$ to the latent embedding $\{\bm{z}^0, \bm{f}_r\}\in\mathbb{R}^{c\times h\times w}$, where $c,h,w$ denotes channels, height and width of the latent embeddings. After diffusion process, the noised $\bm{z}^T$ is further denoised together with $\bm{f}_r$ to generate the clean image $\bm{I}_{O}$. For preserving source-regarding attributes, the novel Region-responsive Semantic Composition is proposed (Sec.~\ref{subsec:RSC}) to extract semantic visual tokens from $\bm{I}_{S}$ by leveraging ArcFace and slot attention, which then control the denoising procedure for better attributes preservation. To train the model, we utilize an attention consistency regularization to inject the prior knowledge embedded in the slot attention, along with the noise prediction objective (Sec.~\ref{subsec: loss}).

\noindent\textbf{Space-sensitive Physical Customization (SPC).} Motivated by the inherent decoupling of physical attributes in 3DMM, we introduce FLAME as the 3D face model and employ off-the-shelf estimator DECA to produce the physical rendering we need. During training, DECA takes query image $\bm{I}_{Q}$ as input and predicts the parameters of FLAME, \ie, shape $\boldsymbol{\beta}_Q$, expression $\boldsymbol{\psi}$ and pose $\boldsymbol{\rho}$, as well as other additional inputs required during rendering including albedo $\boldsymbol{\alpha}_Q$, camera $\boldsymbol{c}$ and Spherical Harmonics lighting $\boldsymbol{l}$, as depicted in Eq.~\ref{eq:3dmm}. During testing, since source and query identities may be different, $\boldsymbol{\alpha}_Q$ and $\boldsymbol{\beta}_Q$ are replaced by their counterpart from $\bm{I}_{S}$, \ie, $\boldsymbol{\alpha}_S$ and $\boldsymbol{\beta}_S$. Subsequently, we can generate corresponding texture rendering $\bm{I}_{R}\in\mathbb{R}^{3\times H\times W}$ as the imposed physical condition using the rendering equation:
\begin{equation}
    \begin{aligned}
        \bm{I}_{R} = \bm{\mathcal{R}}(\boldsymbol{\beta}_i,\boldsymbol{\rho},\boldsymbol{\psi},\boldsymbol{\alpha}_i,\boldsymbol{l},\boldsymbol{c}),\quad i\in\{S,Q\}
    \end{aligned}
\end{equation}
where $\bm{\mathcal{R}}$ denotes the rendering function. $\bm{I}_{R}$ can provide information about edit-required attributes such as pose, expression and lighting, thus well functioning for processing query attributes. In addition to the texture base coefficients, we also leverage the detailed texture information within the UV map. This further enhances high-frequency details in facial regions. Using rendering texture as explicit physical condition indicating pose, expression, and lighting variations, we can decouple and edit these physical attributes to achieve customized effects. 

\subsection{Region-responsive Semantic Composition}
\label{subsec:RSC}

We extract semantic source visual tokens via region-responsive semantic composition.
While the rendering texture $\bm{I}_{R}$ produced by DECA can provide sufficient information for query attribute, since DECA and FLAME cannot concentrate on the fine-grained details for human faces such as hair and wrinkle, such information will not be presented in $\bm{I}_{R}$. Consequently, $\bm{I}_{R}$ suffers from a realistic gap with real images and can hardly be taken as the final prediction.

To solve this problem, we propose the Region-responsive Semantic Composition (RSC) module to get semantically meaningful feature representations of source image by taking inspiration from text-to-image diffusion models such as LDM. In LDM, the text prompts are processed with text encoders, from which each textual token then corresponds to various elements in the generated images, such as different objects and their relationship. Similarly, we propose to extract visual tokens from source image $\bm{I}_{S}$ that are semantically representative for source attributes such as identity, background and clothes. Then these visual tokens can control the generated image respectively, thus avoiding omitting certain attributes during generation. Specifically, RSC consists of two separate modules to extract identity and other tokens.

\noindent\textbf{Identity token extraction.}
The identity information of each individual can be regarded as a unique stylized attribute. Therefore, we calculate the identity token $\bm{f}_{id}$ as the identity embedding extracted by off-the-shelf face recognition model $\bm{\Theta}^{E}$, \ie, $\bm{f}_{id}=\bm{\Theta}^{E}(\bm{I}_{S})$. Specifically, we implement $\bm{\Theta}^{E}$ as the famous ArcFace~\cite{ArcFace} to ensure $\bm{f}_{id}$ is representative enough. This visual token is then injected into the diffusion model using the Adaptive Instance Normalization (AdaIN) technique~\cite{huang2017adain} to enhance ID preservation. Given an intermediate feature map from the $l$-th layer of diffusion model $\bm{f}_{l}$, we have:
\begin{align}
    \operatorname{AdaIN}\left(\bm{f}_{l}, \bm{f}_{id}\right)&=h^{s} \mathrm{IN}\left(\bm{f}_{l}\right)+h^{b}, \\
    h^{s},h^{b} &= \mathrm{MLP}^{1}(\bm{f}_{id}),
\end{align}
where $\mathrm{MLP}$ denotes a single-layer MLP with the SiLU activation~\cite{elfwing2018silu}, IN is the standard instance normalization. 

\noindent \textbf{Semantic token extraction.} Typically, a portrait consists of four components: face, hair, clothes and background. Motivated by the text-to-image generation model, we aim to obtain feature representations that have disentangled and semantically meaningful information for these four components similar to text embeddings. These decoupled features can independently control the corresponding parts of the final generated face images. Specifically, inspired by the slot attention mechanism~\cite{SlotAttention}, we propose the region-responsive encoder $\boldsymbol{\varphi}^{E}$ instantiated as a CNN U-Net, as shown in \cref{fig:difffae}. The source image $\bm{I}_{S}$ is encoded by $\boldsymbol{\varphi}^E$. Then a set of tokens $\bm{F}_{N_{S}}^{0}=\{\bm{f}_i^0\}_{i=1}^{N_S}$ is initialized, where $N_S$ denotes number of tokens. Each $\bm{f}_i^{0}$ performs N-iterative cross attention with $\boldsymbol{\varphi}^{E}(\bm{I}_{S})$. Ideally, refined $\bm{F}_{N_{S}}^{N}=\{\bm{f}_i^N\}_{i=1}^{N_S}$ are expected to represent different semantic areas in $\bm{I}_{S}$.

The region-responsive encoder $\boldsymbol{\varphi}^E$ can be actively incorporated into our DiffFAE framework once pretrained, which will be introduced in Sec.~\ref{subsec: loss}. As shown in \cref{fig:difffae}, the pretrained $\boldsymbol{\varphi}^E$ takes source image $\bm{I}_{S}$ as input to generate the semantic visual tokens $\bm{F}_{N_{S}}^{N}$. These tokens serve as the same role as textual tokens in LDM, \ie, being fused with $\bm{f}_{u}$ via cross-attention:
\begin{equation}
    \begin{aligned}
        \bm{f}_{u} = \operatorname{CrossAttention}(Q(\Tilde{\bm{f}_{u}}), K(\bm{F}_{N_{S}}^{N}), V(\bm{F}_{N_{S}}^{N})),
    \end{aligned}
\end{equation} 
where $Q, K, V$ are learnable linear projections, $\Tilde{\bm{f}_{u}}$ is the intermediate feature map from the denoising U-Net $\epsilon_\theta$.

By leveraging the above module, we expect the semantic tokens can individually control the specific regions of the generated portrait. Specifically, the tokens corresponding to areas including hair, background, and clothes should adaptively influence the generation of these regions respectively based on the information provided by source image $\bm{I}_{S}$, especially when there are extreme changes in pose (which may affect final generated hair, background and clothes to a large extent). Additionally, certain tokens can learn and provide high-frequency facial details, such as teeth and wrinkles, for the generated facial region.

\subsection{Loss Function}
\label{subsec: loss}
\noindent\textbf{Pretraining of RSC.} We follow slot attention to pretrain our region-responsive encoder $\boldsymbol{\varphi}^{E}$ with the reconstruction objective. In concrete, before training our whole framework, $\boldsymbol{\varphi}^{E}$ is composed with an extra decoder $\bm{\zeta}$ and trained to reconstruct the training images in the dataset. For each image, its corresponding semantic tokens are processed with $\boldsymbol{\varphi}^E$, and used as input of $\bm{\zeta}$ to generate the corresponding image and its mask of each token. These images are merged together to rebuild the input image, thus constructing the supervision.

\noindent\textbf{Finetuning of LDM.} To train our model, we follow LDM to utilize the objective $\bm{\mathcal{L}}_{LDM}$ as in Eq.~\ref{eq:ldm} to train the diffusion model jointly with our proposed modules. In addition to the noise prediction objective, we seek a new loss function to enhance the semantic understanding ability of our model, based on the observation that the semantic token extraction module as introduced in Sec.~\ref{subsec:RSC} can provide high-quality semantic parsing prior for the human face images. On the other hand, the semantic visual tokens have to attend on the corresponding region in denoised query feature $\bm{z}_Q^t$ to inject correct information. Therefore, we propose a novel attention consistency regularization (ACR) $\bm{\mathcal{L}}_{attn}$. Specifically, for each source and query image pair $\{\bm{I}_S, \bm{I}_Q\}$, we first achieve their semantic tokens $\bm{F}^{N}_{N_{S},S}, \bm{F}^{N}_{N_{S},Q}$ respectively, along with the corresponding attention mask for the query image $\bm{M}_Q$. Then for each source element in $\bm{F}^{N}_{N_{S},S}$, the cross attention map $\bm{A}_{cross}^t$  with regard to it can be collected, resized and merged from each diffusion model layer. After that we can calculate this constraint as:
\begin{equation}
    \bm{\mathcal{L}}_{attn}=\|\bm{A}_{cross}^t-\bm{M}_Q\|^2.
\end{equation}
The overall loss function is thus defined as:
\begin{equation}
    \begin{aligned}
        \bm{\mathcal{L}} = \bm{\mathcal{L}}_{LDM} + \delta\bm{\mathcal{L}}_{attn},
    \end{aligned}
\end{equation}
where $\delta=0.1$ is empirically found to be helpful. 

\section{Experiments}

\subsection{Settings}
\noindent\textbf{Datasets.}
We train our model on the VoxCeleb1~\cite{VoxCeleb} dataset, which consists of over 100k videos crawled from YouTube and has sufficient variations. Following the preprocessing method proposed in ~\cite{VoxProcess}, we extract face images from original videos. We randomly split the dataset based on ID numbers into training and testing sets in an 8:2 ratio, ensuring there is no overlap in IDs between them.

\noindent\textbf{Metrics.}
We use awareness-related metrics to measure the quality and diversity of the generated images, and attribute-related metrics to measure the accuracy of physical attributes. The main awareness-related metric is Fréchet Inception Distance~\cite{FID} (FID), while the attribute-related metrics are Average Pose Distance~\cite{DPE} (APD), Average Expression Distance~\cite{DPE} (AED), Average Lighting Distance~\cite{DPE} (ALD), and Cosine Similarity of Identity Embedding~\cite{DPE} (CSIM). 

\noindent\textbf{Implementation details.}
The architecture of our diffusion model is based on LDM. During training, we jointly train the region-responsive encoder $\varphi^{E}$ and U-Net $\epsilon_{\theta}$ in an end-to-end manner. Specifically, we use Adam~\cite{Adam} as the optimizer, with a learning rate of 0.0001 and a batch size of 16. The model is optimized for 100k iterations and it takes nearly 34 hours on four A100 GPUs.

\noindent\textbf{Comparison methods.}
Our proposed DiffFAE primarily focuses on editing three attributes: pose, expression, and lighting. Therefore, we compare it with several SOTA methods. Among them, DiffusionRig~\cite{DiffusionRig} stands out as our primary comparative method since it allows simultaneous editing of all three attributes. On the other hand, DPE~\cite{DPE} and StyleHEAT~\cite{StyleHEAT} are limited to modifying pose and expression, while Hou~\etal~\cite{Hou} exclusively modifies lighting. 

\begin{figure*}[tp]
    \centering
    % \vspace{-2.0em}
    \includegraphics[width=1.0\linewidth]{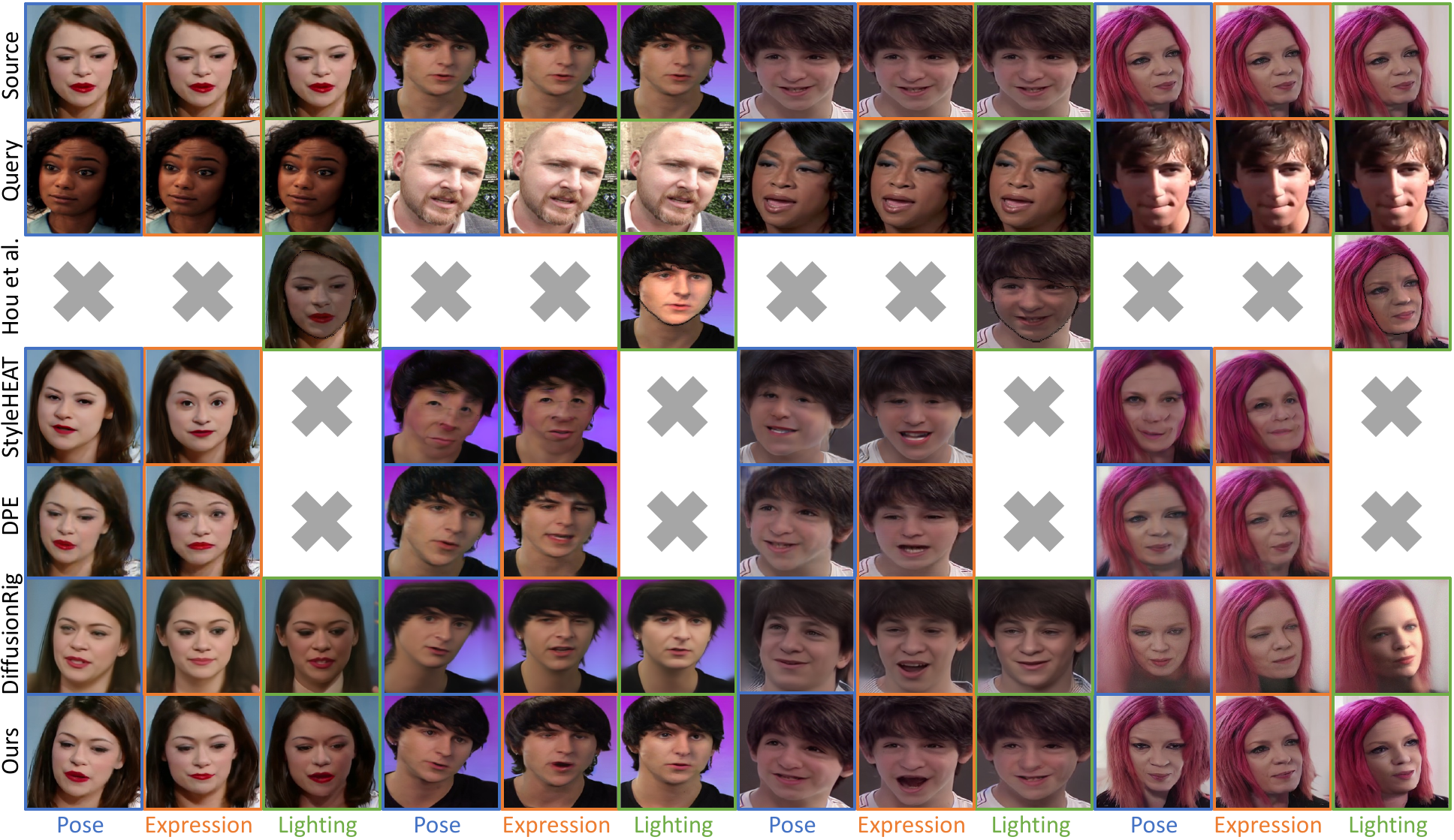}
    \caption{Qualitative comparison between our method and current one-shot SOTAs. Note that Hou~\etal~\cite{Hou}, StyleHEAT~\cite{StyleHEAT} and DPE~\cite{DPE} cannot handle certain types of attributes, hence the corresponding results are net presented.}
    
    \label{fig:qualitative}
    \vspace{-1.0em}
    % \vspace{-0.2in}
\end{figure*}

\subsection{Comparison with SOTAs}

Here we present comparative experiments using facial images in $256 \times 256$ regarding some methods do not support higher resolution, while additional high-resolution and expandable editing results can be found in appendix.

\label{subsec: comparison}
\noindent\textbf{Qualitative results.} We present a comprehensive qualitative comparison in Fig.~\ref{fig:qualitative}. Accordingly, methods in the third to fifth rows can only edit one or two attributes with poor quality. For instance, Hou~\etal does not perform well in terms of changing the lighting, and due to the reliance on facial region mask to maintain background, there are noticeable boundaries between the face and other regions. StyleHEAT does not preserve identity well, resulting in blurry and distorted faces. DPE struggles with extreme pose changes (\eg, from side to front), leading to inaccurate pose attributes in the generated faces. Compared with DiffusionRig which is also of editing all three attributes, our results show better consistency with both the source and query images in terms of the high-frequency details, with no artifacts in facial and non-facial regions. Results show that the general hair style, clothes and background generated by our method keep the same as the source image thanks to the proposed semantic source visual token. Moreover, few-shot finetuning data (roughly 20 personal images) is required for DiffusionRig, while our method only has one training stage without test-time finetuning. This highlights the advantage of our approach in terms of simplicity and efficiency, as it eliminates the requirement for additional data and computational resources during the inference phase.

\begin{table}
    \vspace{-1.0em}
  \caption{Quantitative comparison between our method and the competitors. For FID, APD, AED and ALD, lower score means the better. For CSIM, the higher the better.}
  \centering
  \begin{tabular}{lccccc}
    \toprule
    Method & FID$\downarrow$ & APD$\downarrow$ & AED$\downarrow$ & ALD$\downarrow$ & CSIM$\uparrow$ \\
    \midrule
    Hou~\etal~\cite{Hou} & 49.93 & / & / & 0.1931 & 0.8962 \\
    StyleHEAT~\cite{StyleHEAT} & 78.59 & 0.0425 & 0.1947 & / & 0.8541 \\
    DPE~\cite{DPE} & 36.67 & 0.0490 & 0.1649 & / & 0.9253 \\
    DiffusionRig~\cite{DiffusionRig} & 47.10 & 0.0232 & 0.1727 & \textbf{0.1404} & 0.9196 \\
    Ours & \textbf{31.83} & \textbf{0.0191} & \textbf{0.1532} & 0.1511 & \textbf{0.9458} \\
    \bottomrule
  \end{tabular}
  \label{tab:quantitative}
  \vspace{-1.0em}
\end{table}

\noindent\textbf{Quantitative results.} The quantitative results are presented in Tab.~\ref{tab:quantitative}. Our results are significantly better than the competitors on FID, APD, AED, and CSIM. This demonstrates that our method achieves better performance in terms of accuracy in physical attribute editing, as well as the quality and diversity of the generated images. As for ALD, our method has a gap of 0.0107 with DiffusionRig, which may be attributed to the few-shot samples adopted by DiffusionRig. These samples can help the model better estimate the original lighting information in the source image, thus leading to better results.  However, such finetuning few-shot samples are often hard to access due to privacy problems. Compared with that, our method can make a better balance between data efficiency and performance, showing superiority across multiple evaluation metrics. 

\subsection{Ablation Study}

\begin{figure}[htbp]
	\centering
	\begin{minipage}{0.48\textwidth}
	\centering
        \includegraphics[width=\textwidth]{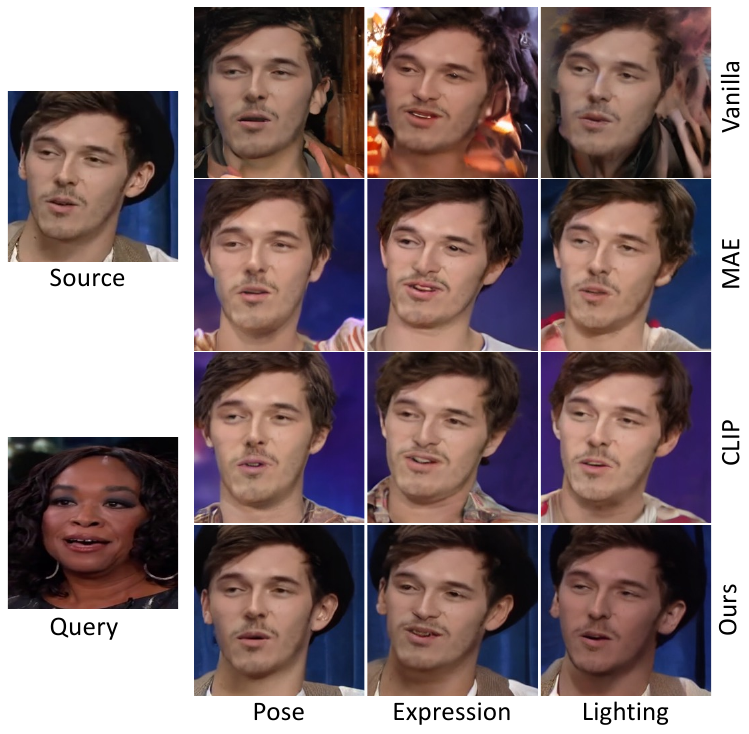}
        \vspace{-2em}
		\caption{Comparison between models with different source image processors.}
		\label{fig:ablation_encoder}
	\end{minipage}
        \quad
	\begin{minipage}{0.48\textwidth}
		\centering
		\includegraphics[width=\textwidth]{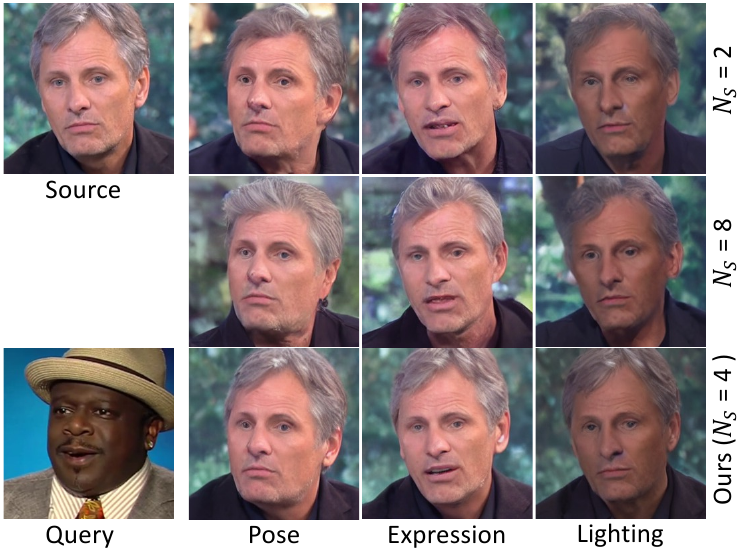}
		\caption{Qualitative comparison between models trained with different number of semantic tokens.}
            \label{fig:ablation_slots_number_qualitative}
	\end{minipage}
\end{figure}
% \vspace{-1.5em}

\noindent\textbf{Effectiveness of semantic tokens.} We test the effectiveness of the proposed semantic token with commonly-used choices for processing conditional images. Concretely, four variants are engaged in this comparison: (1) \textbf{Vanilla}: The VAE encoder of pretrained LDM is adopted. (2) \textbf{MAE}: Due to the powerful encoding capability of MAE~\cite{he2022mae}, we choose this model as one of the baselines. (3) \textbf{CLIP}: Similar to the previous one, the pretrained CLIP image encoder is used. (4) \textbf{Ours}: The semantic tokens are adopted to process the source images. The corresponding results are shown in Fig.~\ref{fig:ablation_encoder}. We can find that the vanilla method can hardly generate regions other than faces, leading to obvious artifacts. Compared with the VAE encoder, MAE and CLIP can help the model generate appropriate images. However, since these methods encode images globally, the information for clothes, hair and background cannot be properly handled, resulting in inconsistency with the source image. Additionally, they lack suitable control when editing lighting conditions. Our proposed semantic tokens, on the other hand, can well deal with all three kinds of physics attributes and other attributes from the source image, thus making the best of both world.

\noindent\textbf{Effectiveness of the number of semantic tokens.} We conduct an ablation study with different token sizes, namely $N_S\in\{2,4,8\}$, to explain how we decide the number of semantic tokens. As shown in Fig.~\ref{fig:ablation_slots_number_qualitative}, when using 2 tokens, the generation results fail to present enough details, which is attributed to insufficient token size. On the other hand, when using 8 tokens, the facial details are nearly desirable. However, this model over-amplifies some non-facial attributes, such as the hairstyle. The reason is that the redundant tokens cannot provide valuable source features, hence they tend to discover dummy information with regard to the source images. We eventually found that when using 4 visual tokens, each token can ultimately represent the face, hairstyle, clothes, and background respectively, providing the most suitable portrait feature representation. Detailed quantitative comparison can be found in \cref{tab:ablation_slots_number_quantitative}.

\begin{figure}[htbp]
	\centering
	\begin{minipage}{0.42\textwidth}
	\centering
        \includegraphics[width=\textwidth]{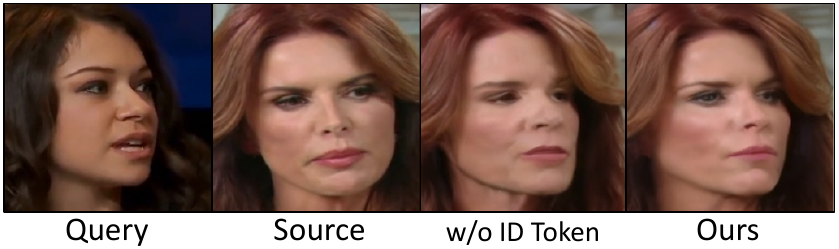}
		\caption{Qualitative comparison (when changing pose) between models trained with and without the identity token.}
		\label{fig:ablation_id_qualitative}
	\end{minipage}
        \quad
	\begin{minipage}{0.54\textwidth}
		\centering
		\includegraphics[width=\textwidth]{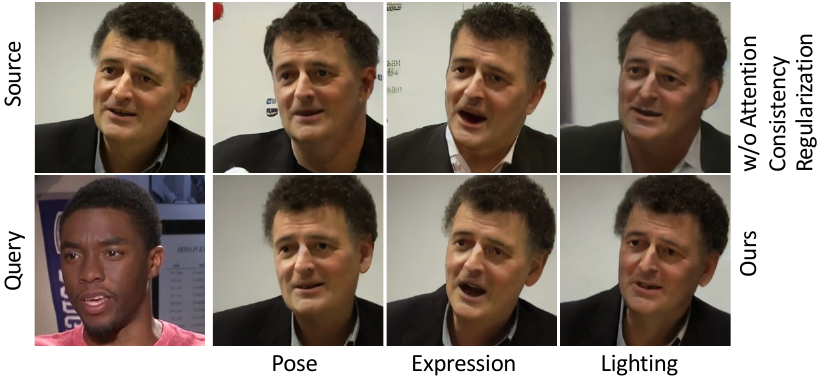}
            \vspace{-2em}
		\caption{Comparison between models trained with and without the proposed attention consistency regularization.}
            \label{fig:ablation_attention_loss}
	\end{minipage}
\end{figure}

\noindent\textbf{Effectiveness of identity token.} We consider identity information as a unique stylized attribute. Therefore, we extract identity token from the source image and employ an AdaIN technique. \cref{tab:ablation-id-quantitative} shows a quantitative enhancement of 0.0149 in the CSIM metric. \cref{fig:ablation_id_qualitative} further indicates that the identity token effectively represents the identity information of the source image, and by injecting it into the denoising model via AdaIN, the capability of id preservation is enhanced.

% \vspace{-0.5em}
\hspace{-1em}
\begin{minipage}[h]{0.5\textwidth}
    \centering
    \captionof{table}{Quantitative comparison with different number of semantic tokens.}
    \begin{tabular}{@{}cccccc@{}}
        \toprule
        $N_{S}$ & FID$\downarrow$ & APD$\downarrow$ & AED$\downarrow$ & ALD$\downarrow$ & CSIM$\uparrow$ \\
        \midrule
        2 & 38.23 & 0.0195 & 0.1572 & 0.1534 & 0.9387 \\
        4 & \textbf{31.83} & \textbf{0.0191} & \textbf{0.1532} & \textbf{0.1511} & \textbf{0.9458} \\
        8 & 35.49 & 0.0198 & 0.1554 & 0.1541 & 0.9404 \\
        \bottomrule
    \end{tabular}  
    \label{tab:ablation_slots_number_quantitative}   
\end{minipage}\qquad
\begin{minipage}[c]{0.4\textwidth}
   \centering
    \captionof{table}{Quantitative comparison between models trained with and without the identity token.}
    % \scalebox{0.9}{
    \begin{tabular}{lc}
        \toprule
        & CSIM $\uparrow$ \\
        \midrule
        w/o identity token &  0.9309\\
        w/ identity token &  \textbf{0.9458} \\
        \bottomrule
    \end{tabular}
    % }
    \label{tab:ablation-id-quantitative}
\end{minipage}
\vspace{+1em}

\noindent\textbf{Effectiveness of attention consistency regularization.} To show the efficacy of the proposed Attention Consistency Regularization (ACR) as in Sec.~\ref{subsec: loss}, we compare the generated results by models trained with and without such an objective in Fig.~\ref{fig:ablation_attention_loss}. The most obvious contrast is that when without ACR, noisy stuff will be added the background. Moreover, it would be hard for the model to keep attributes such as hair and clothes. This is because while the semantic tokens extracted by our method are representative, the LDM is barely trained to understand such type of information during pretraining. Therefore without proper objective to encourage the model make better use of the semantic tokens, their effect will be weakened, leading to worse performance.

% \vspace{-1.5em}
\vspace{-2em}

\subsection{More Discussions}

\begin{figure}[htbp]
	\centering
	\begin{minipage}{0.59\textwidth}
	\centering
        \includegraphics[width=\textwidth]{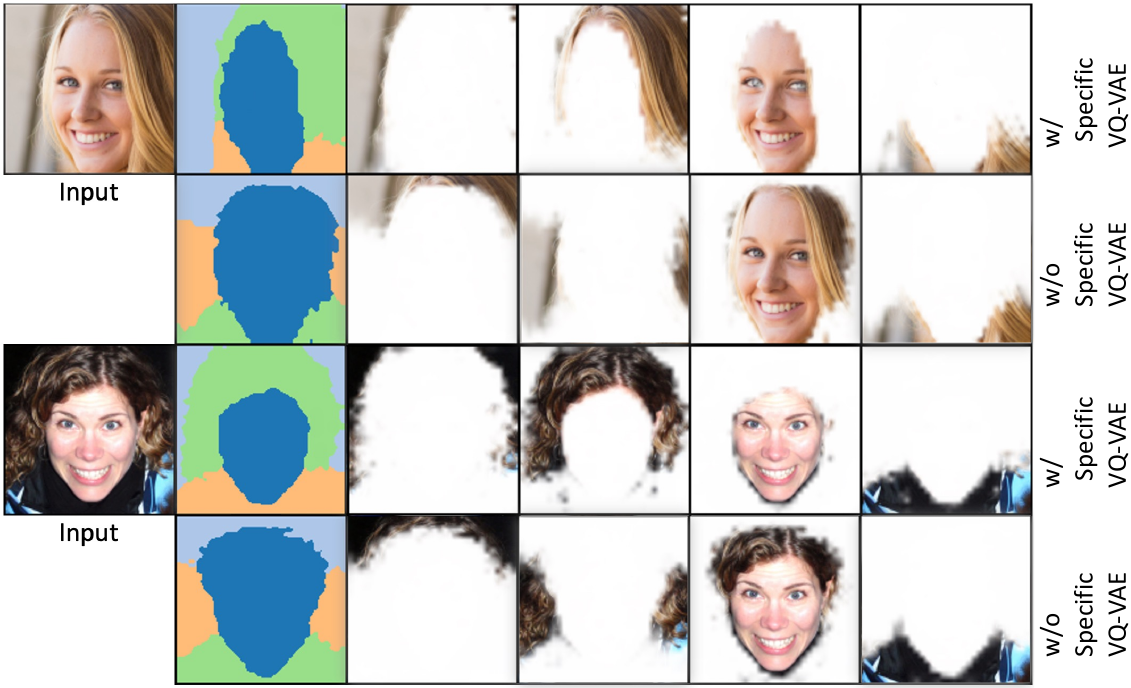}
		\caption{Comparison between region-responsive encoders trained with and without specific VQ-VAE.}
		\label{fig:semantic_mask_vqvae}
	\end{minipage}
        \quad
	\begin{minipage}{0.37\textwidth}
		\centering
		\includegraphics[width=\textwidth]{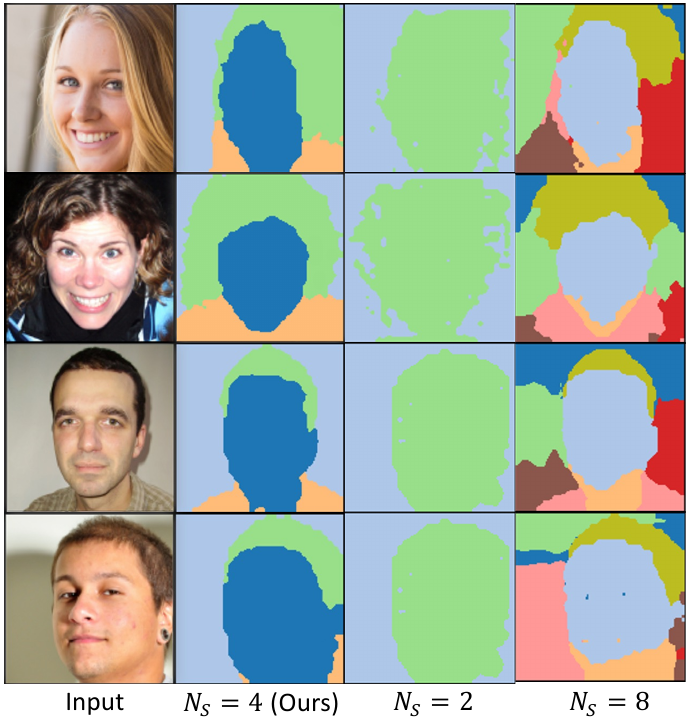}
            \vspace{-2em}
		\caption{Feature representation comparison with different number of semantic tokens.}
		\label{fig:semantic_mask_slots_number}
	\end{minipage}
\end{figure}

\textbf{Impact of VQ-VAE on semantic tokens.} We empirically discover that directly using the VQ-VAE provided by LDM for training the region-responsive encoder does not effectively enable semantic tokens to learn disentangled feature representations. For instance, in the second and fourth rows of ~\cref{fig:semantic_mask_vqvae}, the hair and face regions are coupled together. Consequently, we specifically retrain the VQ-VAE on face images, which significantly improve the representational capacity of the semantic tokens and prevent the coupling of features in different face regions (as seen in the first and third rows of \cref{fig:semantic_mask_vqvae}).

% \vspace{-1.5em}

\noindent\textbf{Impact of different numbers of semantic tokens on semantic prediction.} Here, we additionally demonstrate the impact of varying the number of semantic tokens ($N_{S}$) on the feature representation of portraits. \cref{fig:semantic_mask_slots_number} shows that when $N_{S}$ is set to 2 or 8, the resulting representational capacity is noticeably inferior compared to when $N_{S}=4$, and the corresponding semantic predictions tend to lead to either under-segmentation or over-segmentation, resulting in unfavorable effects. It is because $N_{S}=4$ is the more suitable number of feature representation for portraits, as it explicitly decouples the portraits into four regions: hair, face, background, and clothes. This division into four distinct regions aligns well with the natural structure of a portrait, making it a more intuitive and effective approach. And this arrangement facilitates model learning and strikes a proper balance between token complexity and model's learning ability, enhancing both the semantic prediction and feature representation.

\section{Conclusion}

In this paper we analyse the current challenges in Facial Appearance Editing (FAE), including low generation fidelity, poor attribute preservation and inefficient inference. Based on the analysis we explore a new framework involved by a one-stage diffusion-based pipeline. Specifically, we adopt the Space-sensitive Physical Customization module to deal with the query physical attributes such as pose, expression and lighting. Meanwhile, the Region-responsive Semantic Composition is proposed to better control the source-related attributes. Our method sets new state-of-the-art performance for the FAE task on VoxCeleb1 dataset, which is supported by extensive quantitative and qualitative results.

\noindent\textbf{Limitation and future works.} More powerful 3DMM models can further enhance the model performance, such as better large-pose control and detailed generation. Meanwhile, delving into the realm of temporal Facial Appearance Editing presents an intriguing research avenue for future exploration. 

\bibliographystyle{splncs04}
\bibliography{main}

\clearpage
\begin{center}
    \textbf{\Large Appendix of DiffFAE}
\end{center}
\appendix
% ---------------------------------------------------------------
\section*{Overview}
In this appendix, we present:
\begin{itemize}
    \item \cref{sec:semantic_prediction_comparison}: Semantic Prediction Comparison.
    \item \cref{sec:region_responsive_encoder_details}: Region-responsive Encoder Details.
    \item \cref{sec:impact_of_acr}: Impact of Attention Consistency Regularization.
    \item \cref{sec:network_details}: Neural Network Architecture Details.
    \item \cref{sec:inference_speed_comparison}: Inference Speed Comparison.
    \item \cref{sec:results_high_resolution}: Additional Results (Relighting \& High-resolution Results).
    \item \cref{sec:expandable_editing}: Expandable Editing.
    \item \cref{sec:use_of_dataset}: Responsible Use of Public Dataset.
\end{itemize}

\section{Semantic Prediction Comparison}
\label{sec:semantic_prediction_comparison}

\begin{figure}[tp]
    % \vspace{-2em}
    \centering
	\includegraphics[width=0.7\textwidth]{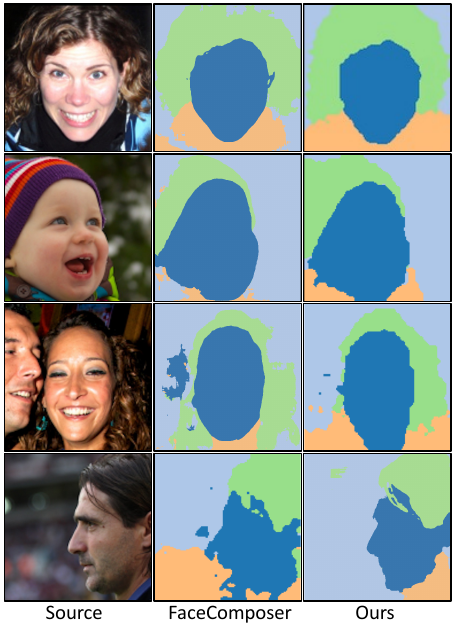}
	\caption{Comparison of semantic prediction.}
    \label{fig:semantic_failure_case}
    % \vspace{-2em}
\end{figure}

Here, we would like to first present some failure cases in semantic prediction. In extreme scenarios, such as multiple faces in the third row and extreme profile face in the fourth row of \cref{fig:semantic_failure_case}, our method may not always achieve accurate prediction, but still exhibit better robustness. Then we would like to highlight that we are the first to explore object-centric learning in human facial editing. Our RSC is fundamentally different from mask-based method like FaceComposer~\cite{FaceComposer}, as we do not require any face parsing model or data. The proposed semantic tokens can learn meaningful semantic regions solely through self-supervised pretraining. To further illustrate this, we compare our method with the face parsing model used in the existing mask-based method~\cite{FaceComposer} in terms of semantic prediction. As can be seen from the first and second rows of \cref{fig:semantic_failure_case}, our method outperforms in most cases, achieving more accurate semantic prediction. This further validates the effectiveness and superiority of the proposed RSC.

\section{Region-responsive Encoder Details}
\label{sec:region_responsive_encoder_details}

The region-responsive encoder $\bm{\varphi}^{E}$ employs a CNN as its foundation, converting raw image into vectors that mesh well with the slot attention mechanism. Specifically, a U-Net framework is embedded within $\bm{\varphi}^{E}$, acting as the pivotal image encoding mechanism. U-Net was selected for its proficiency in auto-encoding, enabling it to weave broad, high-level contextual information throughout the resulting feature sets and slots. Additionally, the inclusion of skip connections within the U-Net architecture ensures that the intricate, low-level details captured by the initial CNN stages are preserved in the final feature output. This approach guarantees that the produced representations are imbued with both detailed texture information and overarching, region-specific data. Details regarding the hyperparameters of $\bm{\varphi}^{E}$ can be found in \cref{tab:hyperparameters_encoder}.

\begin{table}[htp]
  \centering
  \caption{Hyperparameters for our region-responsive encoder.}
  \begin{tabular}{@{}llc@{}}
    \toprule
    CNN Backbone & Input Resolution & 256 $\times$ 256 \\
                 & Output Resolution & 32 $\times$ 32 \\
                 & Self Attention & middle layer \\
                 & Base Channels & 64 \\
                 & Channel Multipliers & 1,1,2,4 \\
                 & Number of Heads & 8 \\
                 & Number of Res Blocks & 2 \\
                 & Output Channels & 192 \\
    \midrule
    Slot Attention & Input Resolution & 32 $\times$ 32 \\
                   & Number of Iterations & 3 \\
                   & Slot Size & 192 \\
                   & Number of Slots & 4 \\
    \bottomrule
  \end{tabular}
  \label{tab:hyperparameters_encoder}
  % \vspace{-2.0em}
\end{table}

\section{Impact of Attention Consistency Regularization} 
\label{sec:impact_of_acr}
\begin{figure*}[htp]\centering
	\includegraphics[width=\linewidth]{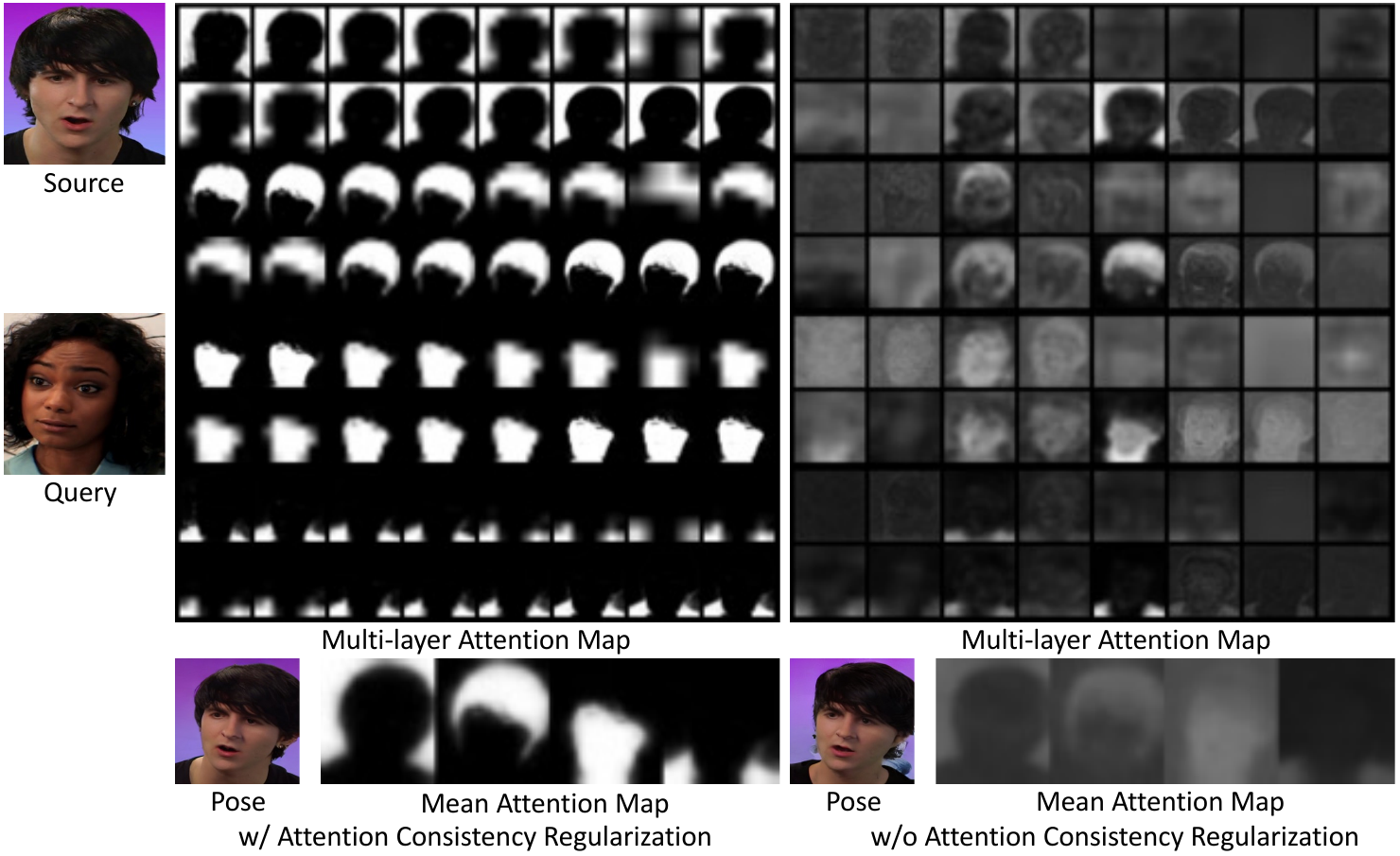}
	\caption{Comparison between models trained with and without the attention consistency regularization.}
    \label{fig:attention_map}
    \vspace{-3em}
\end{figure*}

We also further demonstrate the impact of the proposed attention consistency regularization. \cref{fig:attention_map} shows the attention maps of different layers of the denoising U-Net $\bm{\epsilon}_\theta$, as well as the mean attention map across all layers. It can be observed that without employing attention consistency regularization, the controllability of each semantic tokens over different regions of the final generated face image is significantly weakened, leading to much more artifacts.

\section{Neural Network Architecture Details}
\label{sec:network_details}

Our model is based on the LDM framework, and we have modified the original architecture to facilitate the injection of various conditions as shown in ~\cref{tab:hyperparameters_diffusion}.

\begin{table}[htp]
  \centering
    \caption{Hyperparameters for our latent diffusion model.}
  \begin{tabular}{lcc}
    \toprule
    & 256 $\times$ 256 & 512 $\times$ 512\\
    \midrule
    $z$-shape & 32 $\times$ 32 $\times$ 4 & 64 $\times$ 64 $\times$ 4 \\
    Diffusion Steps &  1000 & 1000 \\
    Noise Schedule &  linear & linear \\
    Channels & 320 & 320 \\
    Depth & 2 & 2 \\
    Channel Multiplier & 1,2,4,4 & 1,2,4,4 \\
    Number of Heads & 8 & 8 \\
    Attention Resolutions & 8,16,32 & 16,32,64 \\
    Embedding Dimension & 192 & 192 \\
    Transformer Depth & 1 & 1 \\
    Batch Size & 16 & 16 \\
    Iterations & 100k & 100k \\
    Learning Rate & 1e-4 & 1e-4 \\
    \bottomrule
  \end{tabular}
  \label{tab:hyperparameters_diffusion}
  \vspace{-0.5em}
\end{table}

\section{Inference Speed Comparison}
\label{sec:inference_speed_comparison}
Here, we present the inference speed comparison. Notably, compared to DiffusionRig~\cite{DiffusionRig}, which is also diffusion-based, our method does not require test-time finetuning and does not inference at pixel level, resulting in a substantial reduction in total time as shown in \cref{tab:inference_speed}. Moreover, our method currently only uses the DDIM~\cite{DDIM} sampler, yet it can still achieve satisfactory inference speed. Faster samplers such as DPM-Solver++~\cite{DPM-Solver++} can be used to speed up inference.

\begin{table}[htp]
    \centering
    \caption{Inference Speed Comparison between our method and DiffusionRig.}
    \begin{tabular}{lcc}
        \toprule
         Method & Finetuning & Inference Speed \\
         \midrule
         % Hou et al.~\cite{Hou} & 0.7269s \\
         % StyleHEAT~\cite{StyleHEAT} & 0.8256s \\
         % DPE~\cite{DPE} & 0.2017s \\
         DiffusionRig~\cite{DiffusionRig} & 21min & 0.8379s \\
         Ours & 0 & 0.8126s \\
         \bottomrule
    \end{tabular}
    \label{tab:inference_speed}
    \vspace{-1.0em}
\end{table}

\section{Additional Results}
\label{sec:results_high_resolution}

\begin{figure}[htp]
    % \vspace{-4em}
    \centering
	\includegraphics[width=\textwidth]{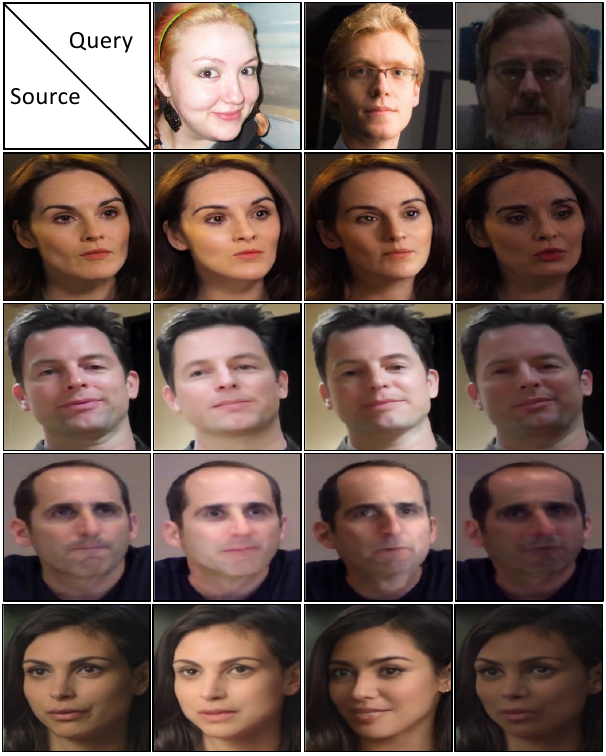}
	\caption{Additional relighting results under various and challenging lighting conditions.}
    \label{fig:relighting_results}
    % \vspace{-5em}
\end{figure}

\begin{figure*}[htp]\centering
	\includegraphics[width=\textwidth]{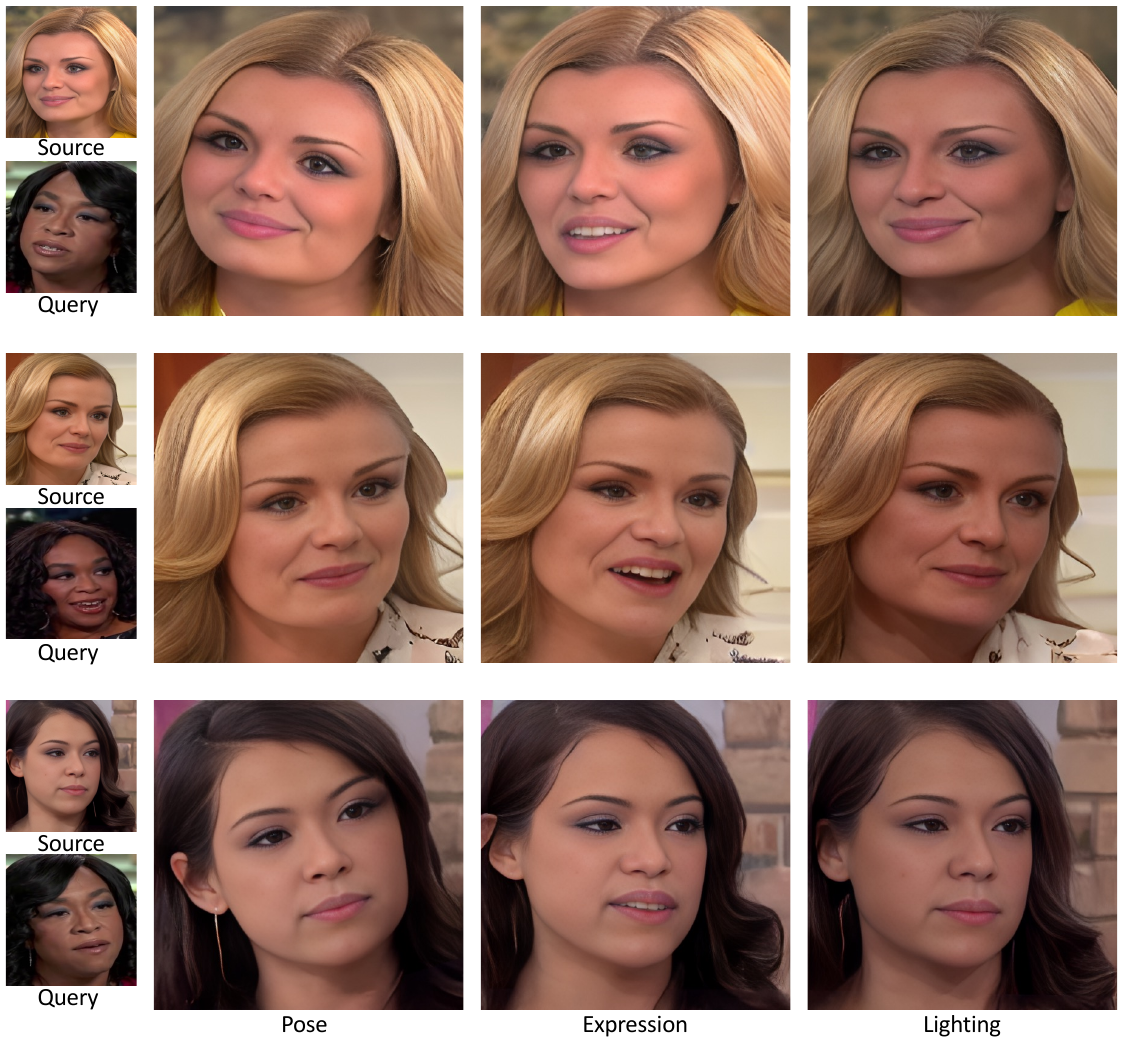}
	\caption{Additional high-resolution facial appearance editing results.}
    \label{fig:hr_results_1}
    % \vspace{-0.2in}
\end{figure*}

\begin{figure*}[htp]\centering
	\includegraphics[width=\linewidth]{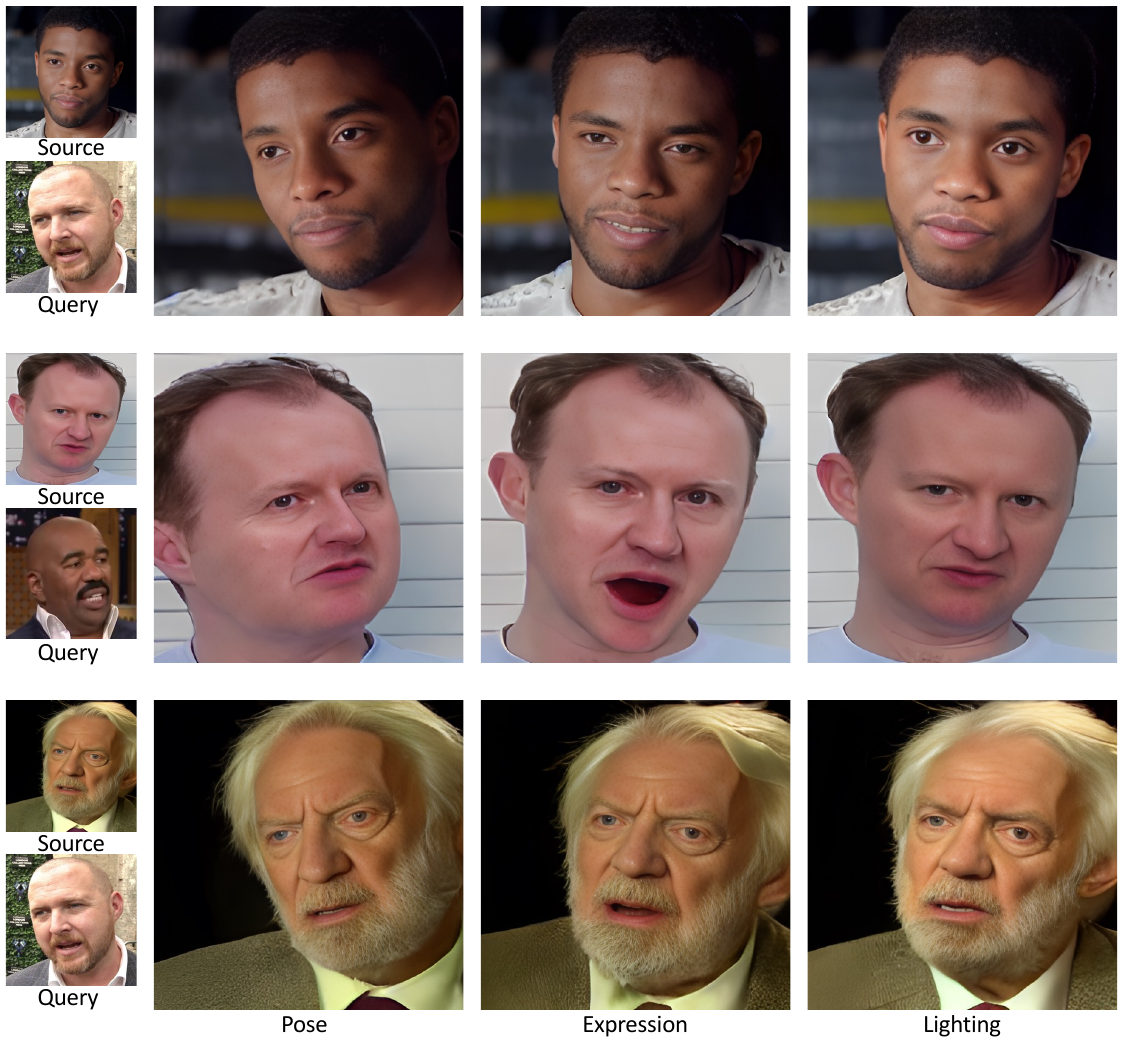}
	\caption{Additional high-resolution facial appearance editing results.}
    \label{fig:hr_results_2}
    % \vspace{-0.2in}
\end{figure*}

Here, we present the results (only changing lighting) under more challenging lighting conditions as shown in \cref{fig:relighting_results}, demonstrating that our method is capable of generating relighting results that adapt well to different lighting conditions. Moreover, our DiffFAE can be effectively scaled up to higher resolutions. ~\cref{fig:hr_results_1} and ~\cref{fig:hr_results_2} show the high-resolution results. 

\section{Expandable Editing}
\label{sec:expandable_editing}
\begin{figure*}[hp]\centering
	\includegraphics[width=\linewidth]{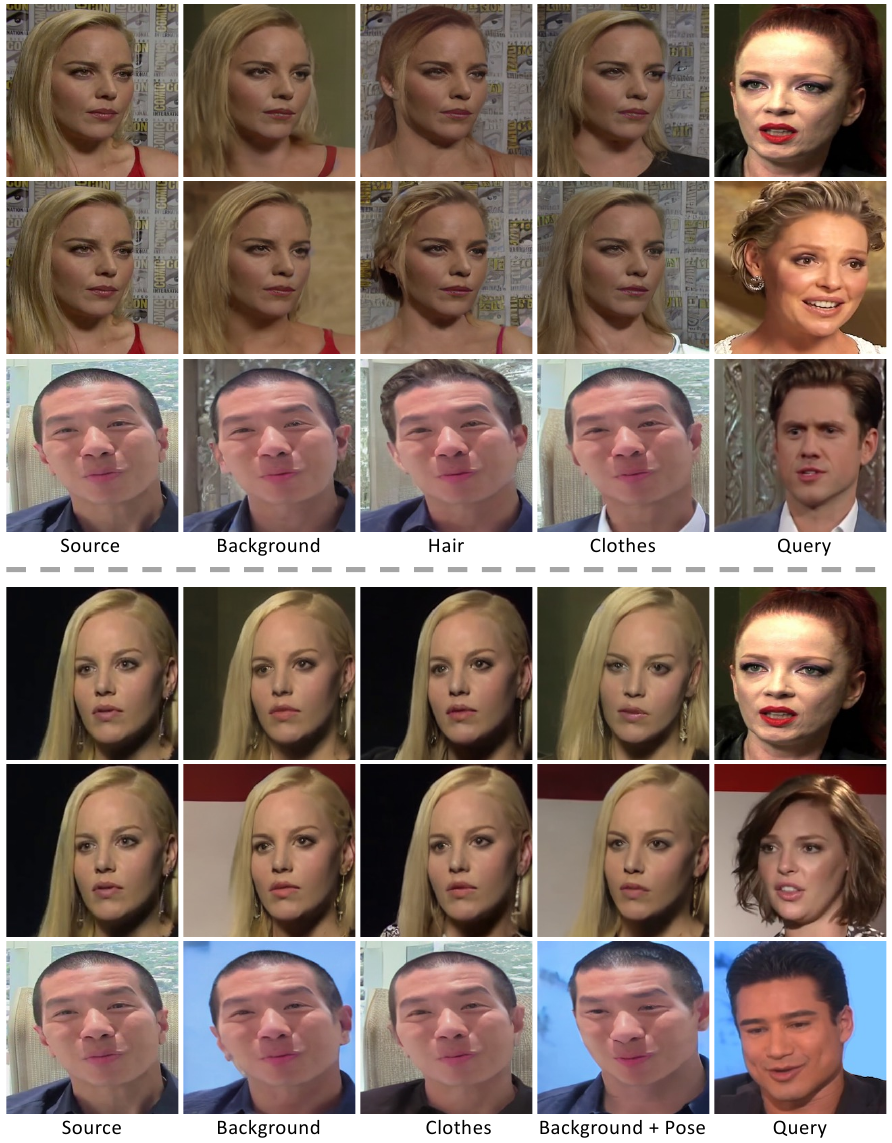}
	\caption{Expandable editing of our DiffFAE. \textbf{Top:} disentangled editing of non-facial attributes (\eg, background, hair and clothes). \textbf{Bottom:} compositional editing of both non-facial attributes and physical properties.}
    \label{fig:expandable_editing}
    % \vspace{-0.2in}
\end{figure*}

Due to the effective disentangled representation of face image achieved by our proposed semantic tokens, our model is capable of editing additional attributes, such as swapping the clothes, hair and background of the portraits. Expandable editing is realized by replacing the source semantic tokens with query ones. \cref{fig:expandable_editing} shows the expandable editing of non-facial attributes as well as the compositional editing that integrates non-facial attributes with physical properties.

\section{Responsible Use of Public Dataset}
\label{sec:use_of_dataset}
We use a public dataset (VoxCeleb\cite{VoxCeleb}), adhering to ethical guidelines and privacy protections. As we do not collect the data, we do not directly obtain consent, but ensure responsible use. The dataset is anonymized, containing no personally identifiable or offensive content. 

\end{document}